\documentclass[runningheads, 10pt, letterpaper]{llncs}

\usepackage{geometry, color}
\input{mysymbol.sty}
\usepackage{algorithm} 
\usepackage{algpseudocode} 
\usepackage{amssymb}
\usepackage{amsmath}
\usepackage[hidelinks]{hyperref}
\usepackage{graphicx}
\usepackage{wrapfig}
\usepackage[numbers]{natbib}
\usepackage{appendix}

\newcommand\blfootnote[1]{%
  \begingroup
  \renewcommand\thefootnote{}\footnote{#1}%
  \addtocounter{footnote}{-1}%
  \endgroup
}

\begin{document}
\title{GraSSRep: Graph-Based Self-Supervised Learning for Repeat Detection in Metagenomic Assembly}
%
%
\author{Ali Azizpour\inst{1} \and
Advait Balaji\inst{2} \and
Todd J. Treangen \inst{2} \and
Santiago Segarra\inst{1}}
\authorrunning{A. Azizpour et al.}
%
\institute{Department of Electrical and Computer Engineering, Rice University, Houston, TX, USA \\
\email{\{aa210,segarra\}@rice.edu} \and
Department of Computer Science, Rice University, Houston, TX, USA \\
\email{\{advait,treangen\}@rice.edu}}
\maketitle              
\begin{abstract}

Repetitive DNA (repeats) poses significant challenges for accurate and efficient genome assembly and sequence alignment. This is particularly true for metagenomic data, where genome dynamics such as horizontal gene transfer, gene duplication, and gene loss/gain complicate accurate genome assembly from metagenomic communities. 
Detecting repeats is a crucial first step in overcoming these challenges. 
To address this issue, we propose GraSSRep, a novel approach that leverages the assembly graph's structure through graph neural networks (GNNs) within a self-supervised learning framework to classify DNA sequences into repetitive and non-repetitive categories.
Specifically, we frame this problem as a node classification task within a metagenomic assembly graph.
In a self-supervised fashion, we rely on a high-precision (but low-recall) heuristic to generate pseudo-labels for a small proportion of the nodes.
We then use those pseudo-labels to train a GNN embedding and a random forest classifier to propagate the labels to the remaining nodes. 
In this way, GraSSRep combines sequencing features with pre-defined and learned graph features to achieve state-of-the-art performance in repeat detection.
We evaluate our method using simulated and synthetic metagenomic datasets. 
The results on the simulated data highlight our GraSSRep's robustness to repeat attributes, demonstrating its effectiveness in handling the complexity of repeated sequences. 
Additionally, our experiments with synthetic metagenomic datasets reveal that incorporating the graph structure and the GNN enhances our detection performance.
Finally, in comparative analyses, GraSSRep outperforms existing repeat detection tools with respect to precision and recall.\blfootnote{This work was supported by the NSF under award EF-2126387.}


\end{abstract}
\setcounter{footnote}{0}
\newpage
\section{Introduction}

Metagenomics is a scientific discipline that involves analyzing genetic material obtained from complex uncultured samples housing DNA from diverse organisms~\cite{primer}.
This field utilizes high-throughput sequencing and bioinformatic techniques to characterize and compare the genomic diversity and functional potential of entire microbial communities without the need for isolating and culturing individual organisms~\cite{magreview}. 
The resulting data can provide insights into the ecological roles and evolutionary relationships of the microorganisms present in the sample~\cite{secondgen}.

However, the sequencing of DNA from such samples poses unique challenges.
One of the major challenges in the metagenomic assembly is the presence of repeats~\cite{metaoverview, metassembly}, which are sequences of DNA that are similar or identical to sequences elsewhere in the genome~\cite{repetitive}. 
The challenges posed by repeats in isolated genomes have primarily been addressed through the use of long-read technologies~\cite{longread}. 
However, metagenomics presents a more complex problem as microbial mixtures often contain multiple closely related genomes that differ in just a few locations due to structural variants, such as horizontal gene transfer~\cite{HGT}, gene duplication, and gene loss/gain~\cite{genegain}. Reads spanning the length of individual strains are required to fully resolve these genome-scale repeats present in microbiomes.

These repetitive elements, while natural and abundant in genomes, complicate the process of genome assembly and comparison~\cite{genesis}. 
They intricately tangle the assembly graph, making it difficult to distinguish the order, orientation, and copy number variation of genomes comprising the microbiome under study, resulting in fragmented assemblies.
Moreover, repeats introduce ambiguities for comparative genomics, hindering differentiation between identical or similar regions and complicating the understanding of gene functions, regulatory elements, and their role in genetic disorders~\cite{repetitive}.
To overcome these obstacles, precise identification and annotation of repeated sequences is necessary. 
Unraveling the complexities of repeated sequences is not only crucial for enhancing genome assembly but also essential for deciphering intricate regulatory mechanisms and evolutionary processes.
Indeed, identifying these repeats is foundational for understanding genome stability, gene expression, and disease susceptibility, making the development of accurate repeat detection methods vital for advancing genomic research~\cite{RED}.

\begin{wrapfigure}{r}{0.35\textwidth}
\centering
    \includegraphics[width=0.33\textwidth]{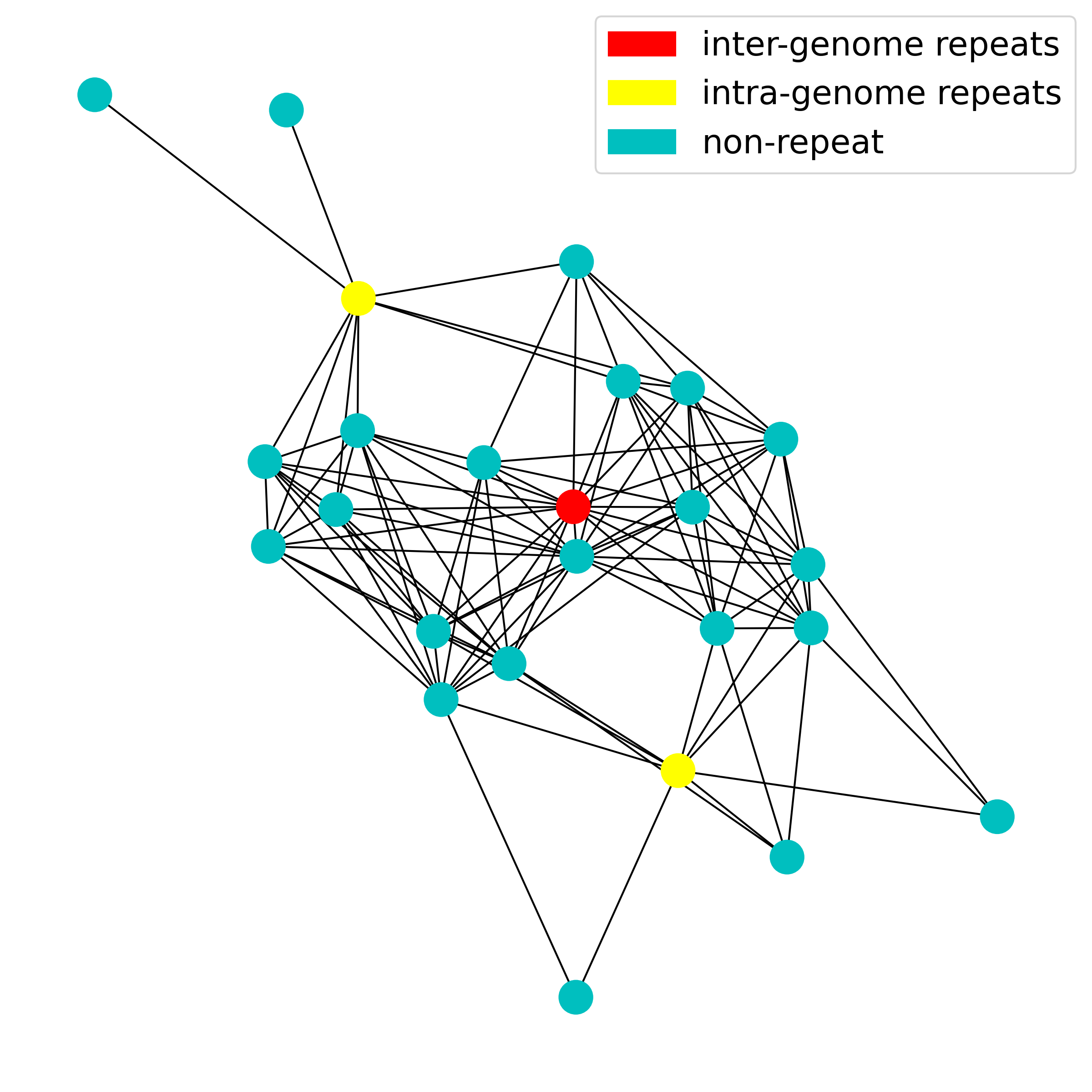}
    \vspace{-5mm}
  \caption{Repeat positions within the assembly graph.}
  \label{simpleGraph}
\end{wrapfigure}

Graphs are powerful tools for visualizing complex relationships between various objects, such as DNA sequences.
Graph-based algorithms can effectively represent the interconnections and overlapping patterns within genomes~\cite{bionetwork}, where the nodes in the graph represent unique DNA sequences. 
Due to the tangled nature of repeated sequences within the assembly graph, exploiting graph structure becomes particularly advantageous.  
As an illustrative example, Figure \ref{simpleGraph} portrays the assembly graph obtained from a simulated metagenome with two organisms.
In this scenario, three random sequences are generated. 
Two of these sequences are inserted as intra-genome repeats in each organism, while the third one is inserted in both organisms, serving as an inter-genome repeat.
The graph reveals that repeats (especially inter-genome repeats) are represented by central and well-connected nodes, indicating the potential of utilizing the inherent graph structure in genomic data for identifying repeated sequences.
{A node labeled as a repeat represents a unique DNA sequence that occurs in several positions of the metagenomic sample.}
Notice, however, that graph structure is not enough to tell apart the yellow nodes from some of the blue ones.
This motivates an approach that combines graph features with sequencing information such as read coverage or length of the DNA sequence.

Previous studies have employed pre-specified graph features in combination with machine learning techniques to address the challenge of detecting repeats, treating it as a node classification problem~~\cite{bi,metacarvel}.
In this context, the nodes of the graph represent DNA sequences, and the objective is to classify them into repeats and non-repeats.
However, given the vast amount of genomic data, there remains ample opportunity for enhancement through learning discriminative graph features. 
One of the promising ways to achieve this is by employing graph neural networks (GNNs)~\cite{gnnsurvey}. 
GNNs have the unique ability to learn distinctive and valuable features for the nodes within the graphs. 
Unlike predefined features, GNNs generate these characteristics through trainable iterative computations, making them adaptive to the specific data. 
These features have shown promising results in many other fields~\cite{otherfields1,otherfields2,otherfields3,otherfields4}, emphasizing the efficiency of utilizing GNNs to classify nodes accurately and uncover the complexities within large graphs~\cite{Representation}.

However, one of the primary challenges in genomic data analysis is the fact that most of the data is unlabeled, particularly in distinguishing between repeat and non-repeat sequences.
This characteristic of the data prevents the application of supervised or semi-supervised learning techniques for classifying DNA sequences~\cite{semisupervised}.
In the absence of labeled data points offering insights into each class, these conventional methods become ineffective.
To overcome this issue, self-supervised learning emerges as a natural and powerful alternative to leverage the vast unsupervised data~\cite{ssl}.
In self-supervised learning, specific data points (nodes) are initially given (potentially noisy) labels.
Subsequently, machine learning algorithms are employed, coupled with fine-tuning steps, to refine the model's performance.
This approach ensures the ability to classify data points without requiring access to their true labels.

In this paper, we propose GraSSRep, a novel graph-based algorithm to identify and detect the repeated sequences in the metagenomic assembly in a self-supervised manner. 
Our contributions are threefold:

\vspace{1mm}
\noindent 
1) By leveraging GNNs, we devise the first method that learns (rather than pre-specifies) graph features for repeat detection.

\vspace{0.5mm}
\noindent 
2) We establish the first algorithm that uses self-supervised learning for repeat detection. In this way, we leverage existing methods to generate noisy labels that we then refine and expand using our learnable architecture.

\vspace{0.5mm}
\noindent 
3) Through numerical experiments, we demonstrate the robustness of our methodology, the value of its constituent parts, and the performance gain compared with the state of the art.

\vspace{1mm}
\noindent 
An implementation of GraSSRep, along with code to reproduce our results in Section~\ref{sec_results}, can be found at {\href{https://github.com/aliaaz99/GraSSRep}{https://github.com/aliaaz99/GraSSRep}}.

\section{Methods}

\begin{figure}[!th]
    \centering
    \includegraphics[width=0.8\textwidth]{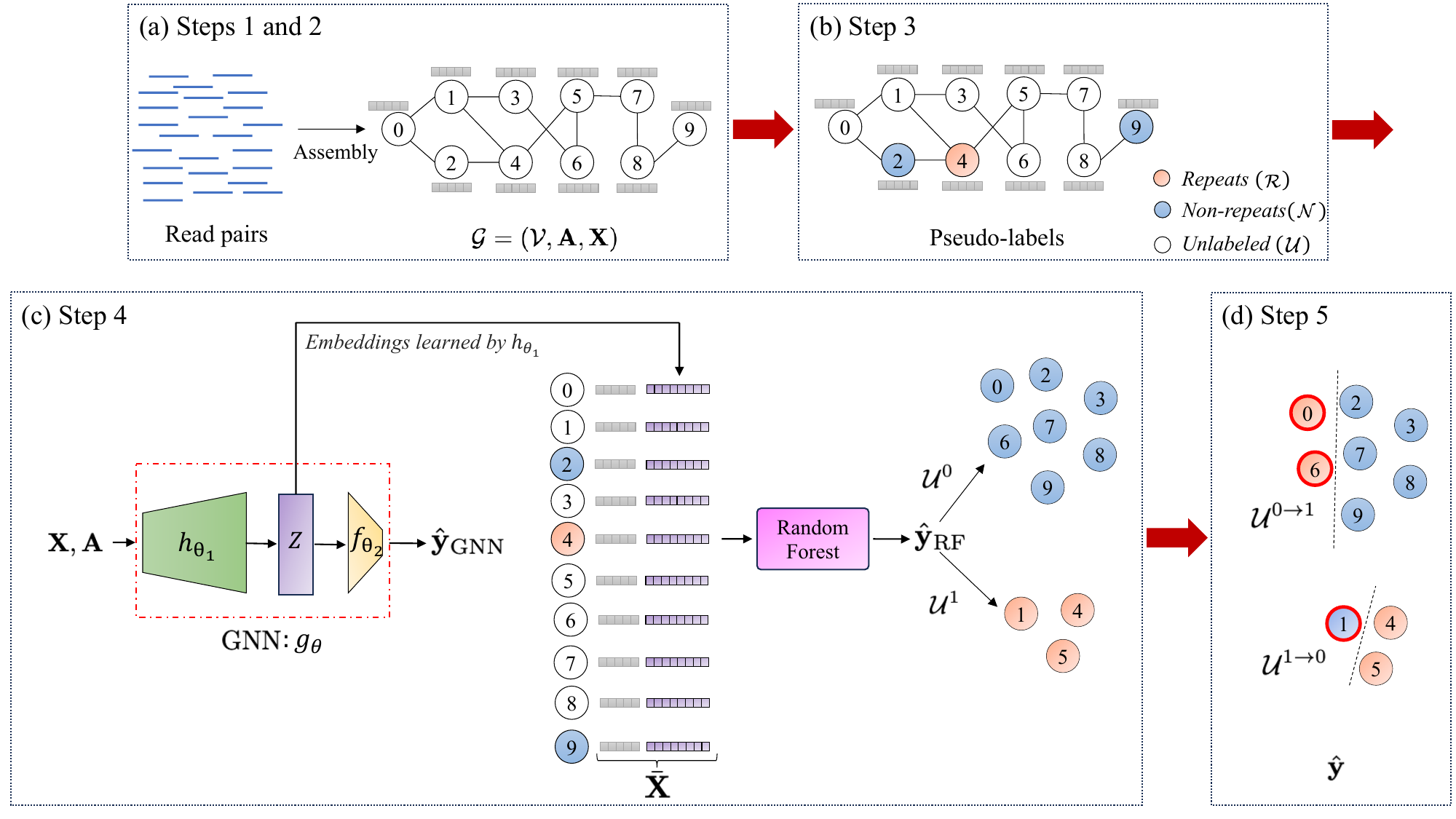}
    \caption{Overview of GraSSRep. 
    (a)~Reads are assembled into unitigs forming the nodes of the unitig graph. 
    Edges are constructed based on the read mapping information. 
    Also, feature vectors are computed for each unitig. 
    (b)~Unitigs with distinctive sequencing features are selected as training nodes and labeled. 
    (c)~The unitig graph is input into a GNN. 
    Embeddings are generated for each unitig and combined with the initial features. 
    A random forest classifier predicts labels for all unitigs based on the augmented feature vectors. 
    (d)~Sequencing features are employed to identify outliers within each predicted class, leading to the reassignment of their class labels.
    \label{fig:overview}
    }
\end{figure}

Given paired-end reads, our goal is to identify repeated DNA sequences in the metagenome (see Section~\ref{ssec_assembly} for the precise criteria used to define a repeat).
An overview of our method specifically designed for this task is illustrated in Figure~\ref{fig:overview}. 
In the subsequent sections, we provide a detailed explanation of each step involved in the pipeline.

\subsection{Step 1: Unitig graph construction}

In the initial step, we construct a unitig graph in order to leverage graph features for repeat detection.
Inspired by KOMB~\cite{komb}, we generate our own graph instead of using existing assemblers. 
We do this due to the tendency of most assemblers to simplify assembly graphs for improved assembly quality, potentially leading to the loss of some sequences. 
Our approach retains all sequences on the graph, facilitating the detection of potential repeats.

First, all reads are assembled into unitigs by feeding them into the de Bruijn graph constructor ABySS~\cite{abyss}. 
Unitigs are obtained by traversing the de Bruijn graph, and the maximal consensus sequences that terminate at branches caused by repeats or variants are regarded as unitigs. 
We consider these unitigs as the nodes $\ccalV$ of our unitig graph, where $|\ccalV| = N$.

To form the edges, we map the reads back to the unitigs using Bowtie2~\cite{bowtie}. 
Based on this, we define {two types} of edges between the unitigs using the information from the read-mapping process. 
The first set, referred to as adjacency edges, captures potentially neighboring unitigs in the genome. 
The second set, which we call repeat edges, provides additional relational information for repeat detection.
Specifically, for a given forward and reverse read pair, denoted as $r_1$ and $r_2$ respectively, where $r_1$ is mapped to a set of unitigs called $U_1$, and $r_2$ is mapped to a set of unitigs called $U_2$, we connect all unitigs in $U_1$ to all unitigs in $U_2$ using adjacency edges. 
Additionally, we establish connections between all unitigs within each set of $U_1$ and $U_2$ using repeat edges.
Furthermore, we assign weights to the edges, given by the number of reads mapping to the corresponding unitigs.
We merge these two edge types to create a single edge connecting the unitigs in our graph. 
The sum of the weights from both repeat and adjacency edges determines the final edge weight.
We then delete the edges with weights in the lowest quartile to focus only on those relations between unitigs that have a strong presence in the reads.
For the remainder of our method, we ignore the weights of the non-deleted edges so that we work with an unweighted graph (except for the computation of the weighted degree in Step 2).
We denote by $\bbA \in \{0,1\}^{N \times N}$ the adjacency matrix of the corresponding unweighted graph, where $A_{ij}=1$ if there is an edge between unitigs $i$ and $j$, and $A_{ij}=0$ otherwise.

\subsection{Step 2: Feature extraction}

We compute features of the unitigs that are informative in determining which unitigs are repetitive.
We consider two types of features: sequencing and graph-based. 
Sequencing features (unitig length and mean coverage) are obtained during the sequencing process before constructing the unitig graph and used in Steps 3 and 5. 
In addition, we incorporate five graph-based features that are widely used in the literature: betweenness centrality, k-core value, degree, weighted degree, and clustering coefficient. 
Previous studies have emphasized the significance of betweenness centrality~\cite{stability_bet} in identifying repeats~\cite{bi}. Additionally, KOMB~\cite{komb} has underscored the crucial role of the k-core value in anomaly detection within unitigs. Furthermore, the simple and weighted degree of nodes indicates their connectivity strength with other unitigs, aiding in the identification of repeated regions. 
We also consider the clustering coefficient due to its substantial impact on node classification tasks, as well as its demonstrated positive effects and favorable outcomes in various related domains~\cite{clusteringcoeff}.
We store the graph-based features in a matrix $\bbX \in \reals^{N \times 5}$, where every row contains the five graph-based features of a given node (unitig) in the graph.
Thus, we define our featured graph of interest $\ccalG = (\ccalV, \bbA, \bbX)$.

\subsection{Step 3: Selection of the training nodes}
\label{ssec:sel_train_node}

Recall that we do not have any prior information (labels) on whether any unitig is a repeat or not.
In this context, the idea of self-supervised learning is first to do a high-confidence classification of a subset of the unitigs (assigning potentially noisy labels, denominated pseudo-labels, to a subset of the nodes) and then use those nodes as a training set for a machine learning model that can classify the remaining unitigs.
We generate this set of pseudo-labels using the sequencing features from Step 2.
In generating pseudo-labels, it is important only to consider those for which we have a high level of confidence, so that the training process based on these pseudo-labels is reliable.

In defining our pseudo-labels, we rely on the fact that shorter unitigs with higher coverage are highly likely to be repetitive, while very long unitigs with lower coverage are more likely to be non-repeat unitigs~\cite{bi}. 
More precisely, let us define as $x^{\mathrm{len}}_i$ and $x^{\mathrm{cov}}_i$ the length (number of base pairs) and coverage (mean number of reads mapped to the base pairs in the unitig) of node $i$, respectively.
We set a percentile $p$ (with $0 \leq p \leq 50$) based on which we define the following thresholds: $\tau^{\mathrm{len}}_{\mathrm{low}}$ is the $p$-th percentile of the lengths among all unitigs in $\ccalV$, $\tau^{\mathrm{len}}_{\mathrm{high}}$ is the $(100-p)$-th percentile of the lengths among all unitigs, and $\tau^{\mathrm{cov}}$ is the $(100-p)$-th percentile of the coverages among all unitigs.
Based on these thresholds, we divide the unitigs into three sets, the repeats $\ccalR$, the non-repeats $\ccalN$, and the unlabeled $\ccalU$, as follows
\begin{align}\label{eq:pseudolabels}
    \ccalR = \{i \in \ccalV \, | \, x^{\mathrm{len}}_i < \tau^{\mathrm{len}}_{\mathrm{low}} \, \land\,   x^{\mathrm{cov}}_i > \tau^{\mathrm{cov}} \}, \qquad 
    \ccalN = \{i \in \ccalV \, | \, x^{\mathrm{len}}_i > \tau^{\mathrm{len}}_{\mathrm{high}} \, \land\,   x^{\mathrm{cov}}_i < \tau^{\mathrm{cov}} \}, 
\end{align}
and $\ccalU = \ccalV \setminus (\ccalR \cup \ccalN)$.
In~\eqref{eq:pseudolabels}, unitigs shorter than the lower length threshold and with a coverage surpassing the coverage threshold are included in the training set with a repeat pseudo-label ($\ccalR$). 
Conversely, unitigs exceeding the higher length threshold and having a coverage below the coverage threshold are added to the training set with a non-repeat pseudo-label ($\ccalN$). 
If a unitig does not meet any of these conditions, it suggests that sequencing features alone are not sufficient to determine its classification. 
Consequently, these unitigs are not included in the training set ($\ccalU$). 

\subsection{Step 4: Unitig classification via self-supervised learning}

We leverage self-supervised learning by training a graph-based model on $\ccalR$ (binary label of $1$) and $\ccalN$ (binary label of $0$) and use that model to classify the nodes in $\ccalU$.

Consider the graph $\ccalG = (\ccalV, \bbA, \bbX)$ generated in Steps 1 and 2 and denote by $g_\theta$ a GNN parameterized by $\theta$~\cite{gnnsurvey}. 
This GNN takes the graph structure $\bbA$ and the node features $\bbX$ as input and produces labels $\hat{\bby}_{\mathrm{GNN}}$ for the nodes at the output. 
To generate these labels, $g_\theta$ can be viewed as an end-to-end network that is structured as follows
\begin{equation}
    \hat{\bby}_{\mathrm{GNN}} = g_\theta(\bbX, \bbA) = f_{\theta_2}(h_{\theta_1}(\bbX, \bbA)),
\end{equation}
where $h_{\theta_1}$ consists of graph convolutional layers followed by an activation function~\cite{relu}.\footnote{We provide here a generic functional description of our methodology whereas in Section~\ref{ssec:node_classification} we detail the specific architecture used in the experiments.}
Each layer in $h_{\theta_1}$ generates new observations for every node based on its neighboring nodes. 
These convolutional layers are succeeded by $f_{\theta_2}$, which represents a fully connected neural network~\cite{neuralnet}. 
The purpose of this network is to predict the final label for each node based on the features derived from the last layer of $h_{\theta_1}$. 

We denote the output of the convolutional layers by $\bbZ = h_{\theta_1}(\bbX, \bbA) \in \reals^{N \times d}$, where $d$ is a pre-specified embedding dimension.
The $i$-th row $\bbz_i$ of $\bbZ$ represents new features for unitig $i$, learned in such a way that the final linear layer, $f_{\theta_2}$, can predict the class of the unitigs based on these features. 
These embeddings enable us to achieve our objective of understanding the graph-based characteristics of repeat and non-repeat unitigs. 
Notice that the features in $\bbz_i$ not only depend on graph features of node $i$ but also on the features of its local neighborhood through the aggregation of the trainable convolutional layers in $h_{\theta_1}$.

In order to learn the parameters $\theta = \{\theta_1 \cup \theta_2 \}$, the GNN undergoes an end-to-end training based on the pseudo-labels $\ccalR$ and $\ccalN$ identified in Step 3.
This training process involves minimizing a loss function that compares the predicted labels $\hat{\bby}_{\mathrm{GNN}}$ with the pseudo-labels
\begin{equation}\label{eq:optim_theta}
    \theta^\star = \argmin_{\theta} \; \sum_{i \in \ccalR}  \mathcal{L} ([\hat{\bby}_{\mathrm{GNN}}(\theta)]_i, 1) + \sum_{i \in \ccalN}  \mathcal{L} ([\hat{\bby}_{\mathrm{GNN}}(\theta)]_i, 0), 
\end{equation}
where $\ccalL$ represents a classification loss (such as cross-entropy loss~\cite{cross_entropy}) and we have made explicit the dependence of $\hat{\bby}_{\mathrm{GNN}}$ with $\theta$.
In essence, in~\eqref{eq:optim_theta} we look for the GNN parameters $\theta^\star$ such that the predicted labels for the nodes in $\ccalR$ are closest to $1$ while the predicted labels for the nodes in $\ccalN$ are closest to $0$.
Intuitively, the intermediate embeddings $\bbZ$ obtained using the optimal parameters $\theta^\star$ encode learning-based features relevant for the classification beyond the pre-defined ones in Step 2.
Thus, we construct the augmented feature matrix $\bar{\bbX} = [\bbX, \bbZ] \in \reals^{N \times (5+d)}$ by concatenating the initial graph-based features with those generated by the GNN.

A random forest (RF) classifier is then trained on the pseudo-labels $\ccalR \cup \ccalN$ having the augmented features $\bar{\bbX}$ as input.
The RF is trained by creating multiple decision trees from different subsets of the dataset (a process known as bootstrapping), with each tree using a random subset of features. When making predictions, the individual trees' outputs are combined through majority voting, producing a reliable and precise ensemble model~\cite{rf}.
The RF classifier combines the explanatory power of the original graph-based features $\bbX$ found to be relevant in previous works with the learning-based features $\bbZ$ to generate the predicted labels $\hat{\bby}_{\mathrm{RF}}$.

Notice that the sequencing features $x^\mathrm{len}$ and $x^\mathrm{cov}$ are not used in computing $\hat{\bby}_{\mathrm{RF}}$ other than in the generation of the pseudo-labels.
If we were to include these features as inputs to the RF, then the classifier can simply learn the conditions in~\eqref{eq:pseudolabels} and obtain zero training error by ignoring all the graph features.
This would directly defeat the purpose of our self-supervised framework.
Instead, the current pipeline can distill the graph-based attributes associated with repeats and non-repeats, enabling us to generalize this knowledge to classify other unitigs effectively.

\subsection{Step 5: Fine-tuning the labels}

In the final step of our method, we enhance the performance of our predictions through a fine-tuning process. 
We first assign the pseudo-labels of the training nodes in $\ccalR$ and $\ccalN$ as their final predicted labels. 
Our primary focus is then directed toward the non-training unitigs in $\ccalU$.
These unitigs have been classified by the RF in Step 4 relying solely on their graph-based features and embeddings learned by the GNN. 
At this point, reconsidering sequencing features becomes crucial, as they hold valuable information that can significantly contribute to determining the accurate labels of the unitigs. 

To do so, we divide the unitigs in $\ccalU$ into two disjoint sets: those predicted as repeats (label 1) by $\hat{\bby}_{\mathrm{RF}}$ form the set $\ccalU^1$ and those predicted as non-repeats (label 0) by $\hat{\bby}_{\mathrm{RF}}$ form the set $\ccalU^0$. 
Within each set, our objective is to identify outliers using the sequencing features $x^\mathrm{len}$ and $x^\mathrm{cov}$ and modify their labels accordingly, similar to Step 3. 
Within each set, specific thresholds are computed based on the distribution of sequencing features of the unitigs in that set. 
More precisely, for $\ccalU^1$ we define $\rho^{\mathrm{len}}_\mathrm{high}$ as the $(100-p)$-th percentile of the unitigs' lengths and $\rho^{\mathrm{cov}}_\mathrm{low}$ and the $p$-th percentile of the coverage.
Conversely, for $\ccalU^0$ we define $\rho^{\mathrm{len}}_\mathrm{low}$ as the $p$-th percentile of the unitigs' lengths and $\rho^{\mathrm{cov}}_\mathrm{high}$ and the $(100-p)$-th percentile of the coverage.
Based on these thresholds, we identify outliers based on the following criteria
\begin{align}\label{eq:outliers}
    \ccalU^{1 \to 0} = \{ i \in \ccalU^1 \, | \,  x^\mathrm{len}_i > \rho^{\mathrm{len}}_{\mathrm{high}} \,\, \land \,\, x^\mathrm{cov}_i < \rho^{\mathrm{cov}}_{\mathrm{low}} \}, \qquad 
    \ccalU^{0 \to 1} = \{ i \in \ccalU^0 \, | \,  x^\mathrm{len}_i < \rho^{\mathrm{len}}_{\mathrm{low}} \,\, \land \,\, x^\mathrm{cov}_i > \rho^{\mathrm{cov}}_{\mathrm{high}} \}.
\end{align}
In~\eqref{eq:outliers}, we change the label from repeat to non-repeat ($\ccalU^{1 \to 0}$) for those unitigs that are longer than a threshold and have low coverage.
Similarly, we change the label from non-repeat to repeat ($\ccalU^{0 \to 1}$) for short unitigs with high coverage.
Notice that we used the same percentile $p$ to compute the thresholds $\rho$ here as that one used to compute the thresholds $\tau$ in Step 3.
Naturally, we could select a different percentile here, but we use the same one as this shows good empirical results and reduces the number of hyperparameters.

Summarizing, the final labels $\hat{\bby}$ predicted by our model are given by
\begin{align}\label{eq:final}
[\hat{\bby}]_i = 
\begin{cases}
    1 \qquad \text{for all } \,\, i \in \ccalR \, \cup \, (\ccalU^1 \setminus \ccalU^{1 \to 0}) \, \cup \, \ccalU^{0 \to 1}, \\
    0 \qquad \text{for all } \,\, i \in \ccalN \, \cup \, (\ccalU^0 \setminus \ccalU^{0 \to 1}) \, \cup \, \ccalU^{1 \to 0}.
\end{cases}
\end{align}
In~\eqref{eq:final}, we see that the unitigs deemed as repeats ($[\hat{\bby}]_i = 1$) by our method are those i)~assigned a repeat pseudo-label in Step 3 ($\ccalR$), ii)~classified as repeats by our RF in Step 4 and not deemed as outliers in Step 5 ($\ccalU^1 \setminus \ccalU^{1 \to 0}$), or iii) classified as non-repeats in Step 4 but later deemed as outliers in Step 5 ($\ccalU^{0 \to 1}$).
Conversely, unitigs classified as non-repeats are those i)~assigned a non-repeat pseudo-label in Step 3 ($\ccalN$), ii)~classified as non-repeats by our RF in Step 4 and not deemed as outliers in Step 5 ($\ccalU^0 \setminus \ccalU^{0 \to 1}$), or iii) classified as repeats in Step 4 but later deemed as outliers in Step 5 ($\ccalU^{1 \to 0}$).

\section{Experimental setup}

\subsection{Datasets}
We test GraSSRep in three types of datasets.

\textbf{Simulated data:} 
To represent distinct organisms, we generate two random backbone genomes with an equal probability of observing each base. 
Subsequently, a random sequence of length $L$ is generated for each backbone and integrated into the genome with a copy number of $C$, serving as an intra-genome repeat. 
Additionally, an inter-genome repeat of length $L$ is randomly generated and inserted $C$ times in both genomes, representing an inter-genome repeat. 
Unlike the backbone genomes, repeats exhibit a non-uniform distribution of bases, resulting in distinctive characteristics unique to each repeat, setting them apart from the backbone genome.
Consequently, we have two genomes, both containing a repeat content of $2 \times L \times C$ within a fixed length of 5 million base pairs for each organism. 
As a result, the characteristics of the repeats within the genomes can be controlled by adjusting the values of $L$ and $C$.
Finally, simulated reads, each 101 base pairs in length, are generated using wgsim~\cite{wgsim} with default values for error (2\%) and mutation (0.1\%). 

\textbf{Shakya 1:} In this dataset, we analyze the reference genomes of a synthetic metagenome called Shakya, which consists of 64 organisms, including 48 bacteria and 16 archaea~\cite{shakya}. 
Based on these reference genomes, read pairs are generated using wgsim, akin to the previous dataset.
However, unlike the simulated data, all the backbone genomes in this dataset are real organisms, containing intricate repeat patterns that are beyond our control. 
The generated reads are 101 base pairs long with a high coverage $(\simeq 50)$, and are produced without any errors or mutations, in order to identify exact repeats in the data.

\textbf{Shakya 2:} 
Read pairs from the Shakya~\cite{shakya} study were obtained from the European Nucleotide Archive (ENA -- Run:\href{https://www.ebi.ac.uk/ena/browser/view/SRR606249}{SRR606249}), all with a length of 101. 
We have no influence over coverage or read errors in this set of reads, mirroring real-world settings. 
This characteristic enables us to evaluate GraSSRep under realistic scenarios.

\subsection{Assembly} 
\label{ssec_assembly}

In all experiments, unitigs are assembled using a k-mer size of $k=51$. 
During the read-mapping step, all reads are trimmed so that only the first $k$ nucleotides are mapped to the unitigs. 
Given that the shortest unitig has a length of $k$, trimming the end of the reads ensures that unitigs with lengths shorter than the read length can still be mapped by some reads.
Therefore, all unitigs are incorporated into the unitig graph, each connecting to at least one other unitig. 
This ensures that our node classification problem comprehensively considers all unitigs generated during the assembly process.

To assess our model accurately, it is crucial to have the ground truth labels for the unitigs. 
To identify these labels, all unitigs are aligned to the reference genomes using NUCmer~\cite{nucmer} (with the \verb|--maxmatch| option).
Unitigs are marked as repeats if they meet specific criteria. 
Generally, this criterion includes aligning at more than one location with at least 95\% identity and 95\% alignment length. 
However, in the error-free cases, the criterion is aligning at more than one location with 100\% identity and alignment length, which indicates exact repeats through the reference genomes (Shakya 1 dataset).

\subsection{Method design and hyperparameter choices}
\label{ssec:node_classification}
{To select and label the training nodes, a threshold value $p$ ranging between 25 and 35 is employed in Step 3, depending on the presence of noise in the data.
Specifically, $p=35$ in instances where noise is present (simulated data and Shakya 2), ensuring robustness in the presence of data irregularities.
For noiseless cases (Shakya 1), we set $p=25$, leading to a stricter definition of repeat pseudo-labels. 
Previous studies have demonstrated that this choice yields effective repeat detection~\cite{metacarvel}.}

In Step 4, the first component of the GNN, $h_{\theta_1}$, consists of two consecutive GraphSAGE convolutional layers, each followed by a ReLU activation function~\cite{sage}. 
The node representation update in these layers can be mathematically defined as follows:
\[
\bbz_v^{(l+1)} = \mathrm{ReLU} \left(\bbW_k \cdot {\text{Mean}}\left(\left\{
\bbz_u^{(l)}, {\forall} u\in \mathrm{Neigh}(v)\right\}\right), \bbB_k \bbz_v^{(l)}\right), \quad {\forall} v\in \mathcal{V},
\]
where \( \bbz_v^{(l)} \) represents the node embedding of the node \( v \) at layer \( l \), \( \mathrm{Neigh}(v) \) represents the set of neighboring nodes of node \( v \), and \( \text{Mean} \) is an aggregation function that combines the embeddings of neighboring nodes.
Moreover, $\bbB_k$ and $\bbW_k$ represent the linear transformation matrix for the self and neighbor embeddings, respectively. 
In this equation, \(\bbz_v^{(l+1)}\) represents the updated embedding of the node \(v\) at the next layer (\(l+1\)). 
The first and second convolutional layers have 16 and 8 hidden channels, respectively.
This results in $d=8$ new features being generated for each node, represented as $\bbZ \in \reals^{N \times 8}$. 
Since $h_{\theta_1}$ has two graph convolutional layers, the final embeddings combine the features within the 2-hop neighborhoods of each node. 
Additionally, the second component of the GNN, $f_{\theta_2}$, comprises a single fully connected layer that transforms the newly learned features, $\bbZ$, into binary classes using a linear transformation matrix $\bbT \in \reals^{8 \times 2}$. 
The GNN is trained for 2000 epochs, utilizing cross-entropy as the loss function and employing the Adam optimizer~\cite{adam} with a learning rate of $0.01$. 

The RF classifier utilizes 100 trees in the forest to generate its results.
The split criterion for each decision tree is determined using the Gini impurity measure, ensuring the creation of optimal splits at each node.
Finally, to account for the randomness inherent in the training process, both the training and testing steps are repeated for 10 iterations in each case. 
The reported results are averaged across these iterations, providing a robust and reliable evaluation.
As figures of merit, we report the classification accuracy, precision, recall, and F1-score (harmonic mean of precision and recall).

\section{Results}
\label{sec_results}

We present a comprehensive analysis of our algorithm's performance across various settings.

\subsection{{Evaluation on varying repeat characteristics}}

We leverage the simulated dataset to examine the effect of three crucial characteristics that are beyond our control within the real datasets:

\vspace{1mm}
\textbf{A) Length of the repeats.} To measure the impact of repeat length, we fix the copy number of both inserted intra-genome and inter-genome repeats at $C=25$ and vary their length from $L=150$ to $L=1000$ base pairs, leading to a copy content ranging from 0.15\% to 1\% in the reference genomes.

\textbf{B) Copy number of the repeats.} We set the length of the inserted repeats to $L=400$ base pairs and adjust their copy number from $C=20$ to $C=150$, increasing the complexity of the dataset. 
This results in a copy content ranging between 0.32\% and 2.4\% in the reference genomes.

\textbf{C) Coverage.} We generate backbone data by inserting repeats of $L=400$ base pairs in length with a copy number of $C=25$ to have 0.4\% copy content in the reference genomes. 
The number of generated read pairs is varied, ranging from 0.5 to 2.5 million base pairs. 
Consequently, the coverage ranges from 10 to 50, allowing us to analyze the algorithm's performance under different coverage levels.

\vspace{1mm}
These adjustments enable a detailed evaluation of our algorithm's robustness and adaptability across a spectrum of repeat characteristics and coverage scenarios. 
Note that due to errors and mutations in the generated reads, our analysis considers a repeat as having at least 95\% identity over 95\% of the length. Consequently, more than just three unitigs are identified as repeats in this context, each with copy numbers that may differ from the exact number of inserted repeats.

\begin{figure}[!th]
    \centering
    \includegraphics[width=\textwidth]{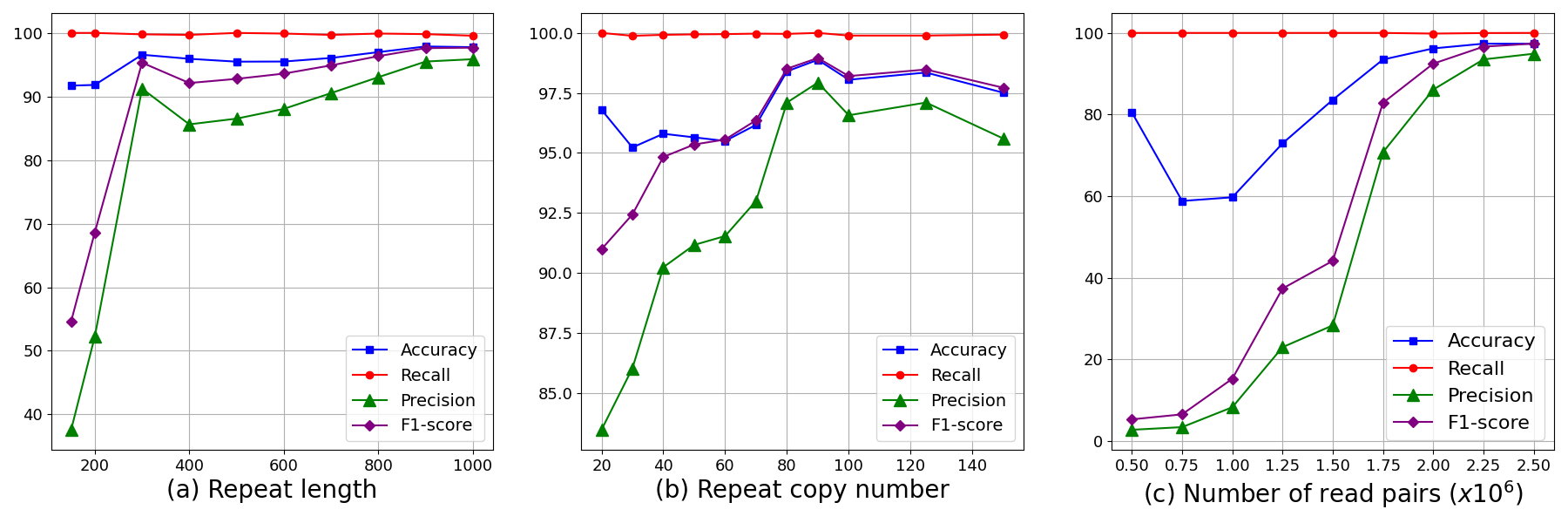}
    \caption{Assessing the method across various repeat characteristics. 
    (a)~The model remains stable in metrics even with increasing repeat length, especially beyond the outer distance of read pairs (500 base pairs). 
    (b)~The method is robust to the copy number variation, consistently achieving an F1-score above 90\%.
    (c)~Higher sequencing coverage improves the model's performance.}
    \label{fig:sim}
\end{figure}

As illustrated in Figure~\ref{fig:sim}(a), our approach demonstrates resilience to variations in repeat length, with all metrics remaining stable as the repeat length increases. 
This robustness is particularly evident when the repeat length surpasses the outer distance of the read pairs, which is 500 in our case. 
This phenomenon occurs because the repeat unitigs predominantly connect with neighboring segments in the reference genome within the graph structure. 
However, when the repeat length falls short of the outer distance, the unitigs corresponding to DNA segments to the left and right of the repeats in the reference genome might be directly connected by an edge in the unitig graph, leading to less precise repeat detection.

Figure~\ref{fig:sim}(b) shows the performance attained when varying the copy number.
Our method achieves an F1-score constantly exceeding 90\% and, with copy numbers greater than 50, the precision also surpasses 90\%.
Given the simulated data scenario, with a 400-length repeat within a 5 million base pairs backbone, it is logical for repeat copy numbers to fall within this range with at most 2\% repeat content. 
This outcome underscores the robustness of GraSSRep to logical copy numbers and demonstrates its resilience not only to repeat length but also to copy number variations. 
This ensures its applicability across various datasets and scenarios.

As demonstrated in Figure~\ref{fig:sim}(c), the model's performance exhibits a constant improvement with increased coverage, particularly in terms of enhanced precision. 
Notably, when the coverage reaches 40 (corresponding to 2 million reads), the model achieves an F1-score greater than 90.
It is important to note that across all the different scenarios in the preceding three cases, our approach consistently achieves an almost perfect recall rate of nearly 100\%, highlighting its effectiveness in detecting almost all repeats.

\subsection{Ablation study of the steps of the algorithm}

We focus on the behavior of our method across different steps using the Shakya 1 dataset. 
After assembling and constructing the graph, we have $N=59808$ unitigs as the nodes of the graph, out of which 11900 unitigs are exact repeats (total length of the unitig repeated with 100\% identity).

To begin, our evaluation involves assessing the method across various steps of the pipeline.
Specifically, we examine the outcomes relative to the baseline, the results produced by the GNN ($\hat{\bby}_{\mathrm{GNN}}$), the outputs generated by RF ($\hat{\bby}_{\mathrm{RF}}$), and finally, after the fine-tuning step ($\hat{\bby}$).
In this context, the term ``baseline" refers to a straightforward heuristic used to classify the unitigs. 
This heuristic relies on Step 3 and labels nodes according to the following criteria
\begin{align}\label{eq:baseline}
[\hat{\bby}_\mathrm{base}]_i = 
\begin{cases}
    1 \qquad \text{for all } \,\, i \in \ccalR, \\
    0 \qquad \text{for all } \,\, i \in \ccalN \, \cup \, \ccalU.
\end{cases}
\end{align}
This approach allows us to test the effectiveness of sequencing features in node labeling in the absence of graph-based features.

\begin{figure}[!th]
    \centering
    \includegraphics[width=\textwidth]{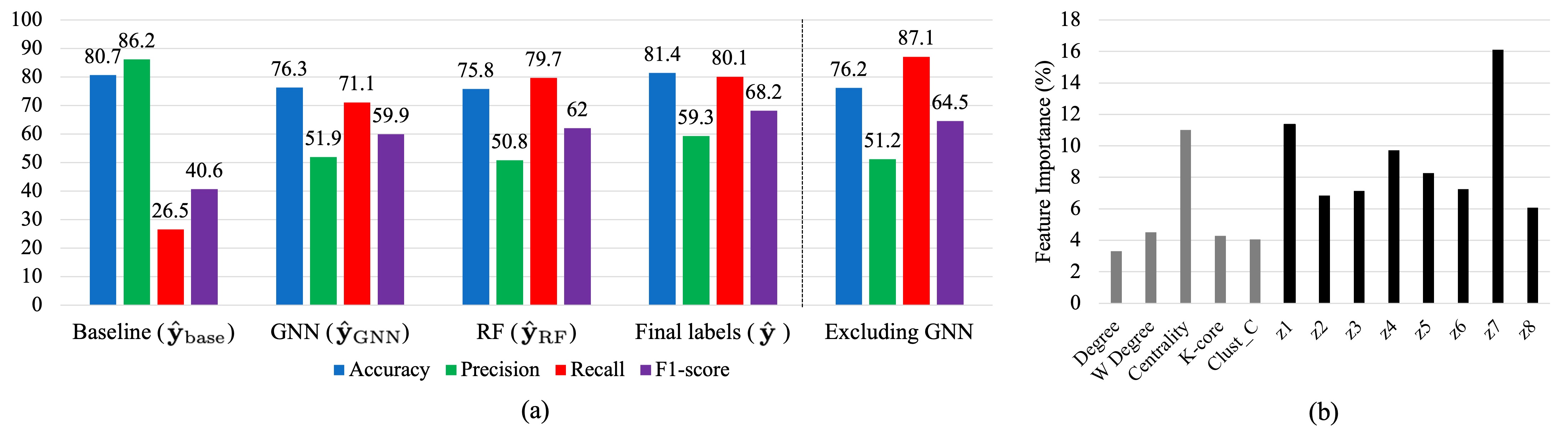}
    \caption{
    (a)~Progression of the method's performance throughout the different steps, highlighting the effectiveness of each step in improving repeat detection. We also test the impact of excluding the GNN embeddings.
    (b)~High importance of GNN-generated embeddings in RF classification. }
    \label{fig:shakya_1_results}
\end{figure}

In Figure~\ref{fig:shakya_1_results}(a), it is evident that the F1-score consistently rises throughout the pipeline, emphasizing the importance of each step in achieving optimal results. 
The baseline method exhibits high precision (86.2\%) but low recall (26.5\%), indicating appropriate node selection for determining pseudo-labels but an inability to identify most repeats.
This observation underscores that sequencing features alone are insufficient for detecting repeats.
This limitation is modified by the GNN, which significantly boosts the recall to 71.1\%, effectively identifying more repeats, which suggests that graph structure is significant in detecting the repeats. 
Subsequent application of the RF further amplifies this increase in recall to 79.7\%. However, this enhanced recall comes at the cost of reduced precision compared to the baseline. 
To address this precision loss, the fine-tuning step effectively identifies outliers, leading to a precision increase from 50.8\% at the output of the RF to 59.3\% for the final estimation. 
In summary, our approach yields a 68.2\% F1-score without any prior labels on the unitigs, representing a substantial improvement of 27.6\% over the baseline method.

Moreover, we investigate the impact of the GNN and the embeddings it generates. 
To assess this, we perform two analyses. 
First, we exclude $\bbZ$ from the feature matrix fed to the RF, resulting in $\bar{\bbX}~=~[\bbX]~\in~\reals^{N \times 5}$  aiming to observe the method's performance only based on the initial graph-based features. 
As depicted in Figure~\ref{fig:shakya_1_results}(a) under `Excluding GNN', this exclusion leads to a decrease in both F1-score and precision. 
This decline suggests that embeddings play a crucial role in enhancing the reliability of repeat detection. 
Second, we calculate the importance of the features fed to the RF by averaging the impurity decrease from each feature across trees. The more a feature decreases the impurity, the more important it is. 
These importance values are then plotted in Figure~\ref{fig:shakya_1_results}(b).
The plot indicates that nearly all learned embeddings (labeled z1 through z8) exhibit higher importance compared to the pre-specified graph features, except for betweenness centrality. 
This finding emphasizes the utility of the embeddings generated by the GNN in improving the overall performance of the method. 

Lastly, we perform an ablation study on the percentile value $p$ used to define the thresholds in Steps 3 and 5. 
The analysis in Appendix~\ref{appendix_A_1} reveals that our approach is robust to this hyperparameter, particularly within the range of 25 to 35, which corresponds to the range used in our experiments.

\subsection{Comparison with existing repeat detection methods}

We present a comprehensive comparison of our method with several existing repeat detection methods using unitigs assembled from the reads downloaded from ENA (Shakya 2). 

\begin{figure}[!th]
    \centering
    \includegraphics[width=\textwidth]{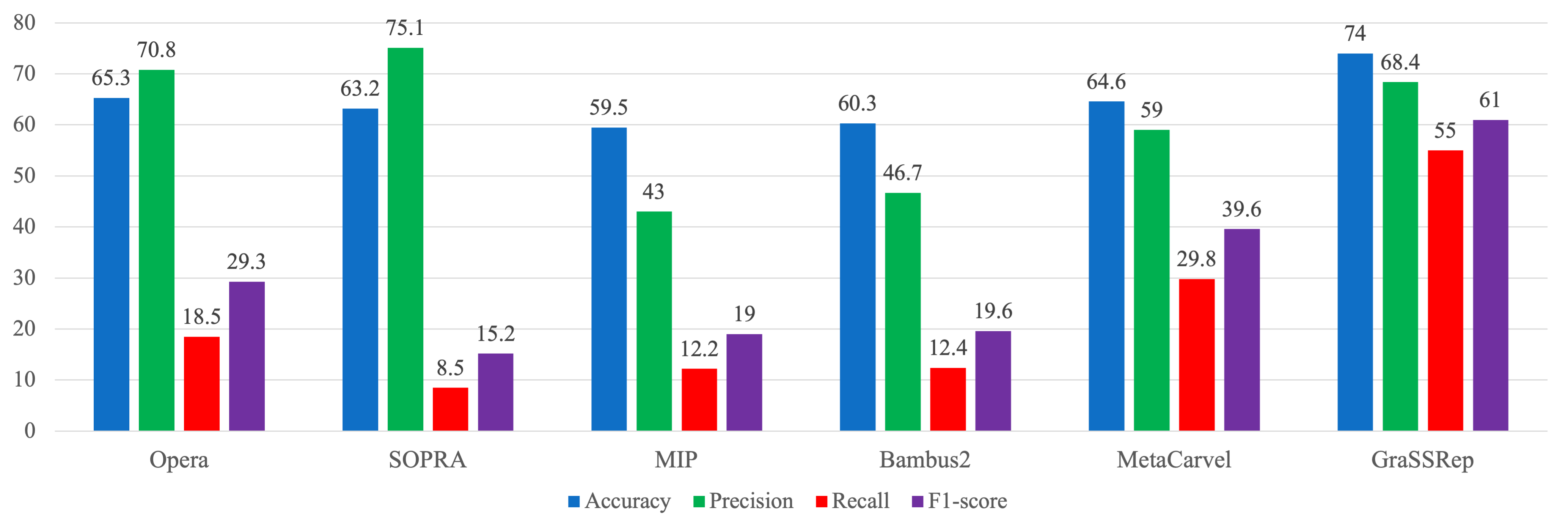}
    \caption{GraSSRep compared to the other repeat detection methods, demonstrating superior performance.}
    \label{fig:shakya_2_1}
\end{figure}

We consider five widely recognized methods for this comparison.
Opera~\cite{opera} and SOPRA~\cite{sopra} identify repetitive unitigs by filtering out those with coverage 1.5 and 2.5 times higher than the average coverage of all unitigs, respectively, without considering any graph structure. 
Similarly, the MIP scaffolder~\cite{mip} utilizes both high coverage (more than 2.5 times the average) and a high degree ($\geq$ 50) within the unitig graph to detect the repeats. 
Additionally, Bambus2~\cite{bambus} categorizes a unitig as a repeat if the betweenness centrality, divided by the unitig length, exceeds the upper bound of the range within $c$ standard deviations above the mean on this feature.
Here, $c$ represents a hyperparameter of this method, and the optimal outcome on our dataset was achieved with $c=0$.
Lastly, Metacarvel~\cite{metacarvel} employs four more complex graph-based features alongside coverage in a two-step process. 
First, any unitig with a high betweenness centrality ($\geq$ three standard deviations plus the mean) on the unitig graph is marked as a repeat.
Moreover, a unitig is identified as a repeat if it falls within the upper quartile for at least three of these features: mean coverage, degree, ratio of skewed edges (based on coverage), and ratio of incident edges invalidated during the orientation phase of the unitigs; see~\cite{metacarvel} for details.
Notably, since we utilize a unitig graph instead of a scaffold graph, we do not incorporate the latest feature and adjust the flag threshold from three to two in the second step.

As illustrated in Figure~\ref{fig:shakya_2_1}, GraSSRep outperforms all other methods, particularly demonstrating superior capability in detecting repeats with a higher recall rate (55\% versus the next best alternative at 29.8\%).
This superiority comes from the combined value of incorporating learnable graph features (through the GNN) and considering a self-supervised framework.
Notice that, even if we fix the embedding dimension at $d=8$, the graph features learned by the GNN depend on the specific dataset under consideration.
In this way, our trainable architecture can distill the key graph features that characterize repeats in the specific metagenomic sample.
This adaptive approach stands in contrast to other methods, which often rely on fixed features.
Moreover, since the RF is not pre-trained but rather trained based on the pseudo-labels, different features may vary in importance based on context.
For instance, if we repeat the analysis in Figure~\ref{fig:shakya_1_results}(b) for this dataset (not shown here), we observe that the clustering coefficient of the unitigs holds greater significance in detecting repeats compared to betweenness centrality or degree.
In this way, our self-supervised framework allows us to adapt to the metagenomic data at hand, and we do not have to worry about generalization issues of pre-trained models.

Thus far, we have focused on the practical unsupervised setting where no repeat labels are available.
For completeness, we now consider the case where repeat labels for some unitigs are available. 
This setting might arise, e.g., if we have knowledge about specific organisms present in the metagenomic sample and their corresponding reference genomes are accessible. 
GraSSRep can seamlessly accommodate this case.
In our pipeline, we can leverage this prior knowledge to substitute Step 3. 
Instead of pseudo-labels, we employ the known node labels as our training set, leading to a semi-supervised (instead of self-supervised) setting.
Our analysis in Appendix~\ref{appendix_A_2} shows that performance can be markedly improved in the case where labels are available for a fraction of the unitigs.

\section{Conclusion}

We tackled the challenging task of detecting repetitive sequences (repeats) in metagenomics data when we only have access to paired-end reads. 
We introduced GraSSRep, a novel method that leverages the inherent structure of the assembly graph by employing GNNs to extract specific features for the unitigs. 
Moreover, adopting a self-supervised learning framework, we generated noisy pseudo-labels for a subset of the unitigs, which were then used to train a graph-based classifier on the rest of the unitigs. 
Experimental studies using both simulated and synthetic metagenomic datasets demonstrated the robustness of our method across diverse repeat characteristics, the value of every step in our algorithm in enhancing repeat detection, and the performance gain compared to existing repeat detection tools.

A natural extension of our approach is its integration into widely used assemblers. 
This integration would replace their existing repeat detection modules with GraSSRep, yielding potential improvements in assembly quality.
We also intend to apply our method to real datasets, particularly in environments like hot springs where widely accessible reference genomes are scarce.
Lastly, the overall pipeline of GraSSRep can potentially address other problems in genomics where graph structures can be used to identify specific genetic markers in the absence of prior knowledge.
For instance, we intend to leverage our approach for the identification of transposable elements, which play important roles in eukaryotic/mammalian genomes. 

\bibliographystyle{splncs04}
\bibliography{ref}

\newpage
\appendix
\section{Appendix}

\subsection{ Ablation study on the percentile value $p$}
\label{appendix_A_1}
\begin{wrapfigure}{r}{0.45\textwidth}
\centering
    \includegraphics[width=0.44\textwidth]{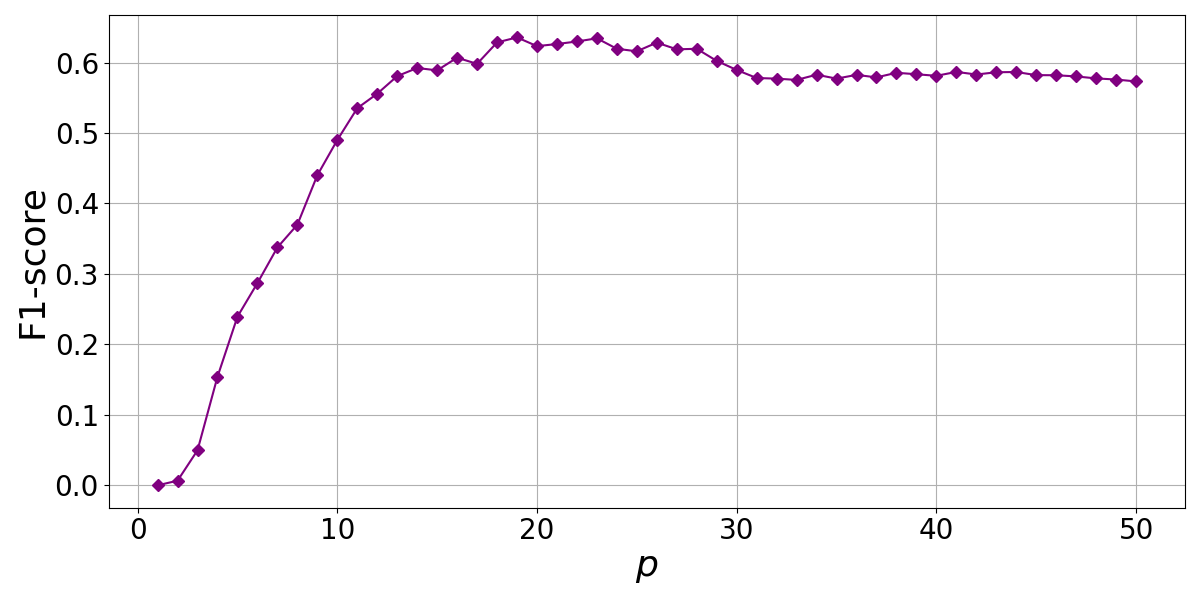}
    \vspace{-5mm}
  \caption{F1-score performance across varying values of the hyperparameter $p$, demonstrating the method's stability.}
  \label{fig:shakya_1_p_appendix}
\end{wrapfigure}

A critical hyperparameter of GraSSRep is the percentile value $p$, which determines the thresholds in Steps 3 and 5. 
In Figure~\ref{fig:shakya_1_p_appendix}, we plot the achieved F1-score for different choices of $p$, ranging from 1 to 50. 

As illustrated, GraSSRep consistently attains an F1-score exceeding 50\% for the values of $p$ greater than 10, indicating the method's robustness to this hyperparameter.

\subsection{Incorporating prior knowledge}
\label{appendix_A_2}
In certain scenarios, prior information about unitigs can be available, allowing for their labeling without relying on sequencing features. 
For instance, if we have knowledge about specific organisms present in the metagenomic sample and their corresponding reference genomes are accessible, we can identify repetitive unitigs associated with those organisms. 
In our pipeline, we leverage this prior knowledge to substitute Step 3. 
Instead, we employ nodes labeled with this prior information as the training set. 
This approach prompted us to explore semi-supervised learning. 

To do this, we initially determine the number of unitigs that would have been included in the training set using Step 3. 
Subsequently, we randomly select the same number of nodes with their labels as our training nodes.
Additionally, instead of fine-tuning the labels predicted by RF with sequencing features in Step 5, we integrate the length and mean coverage of the unitigs into their initial node features within the unitig graph.
The remainder of our method remains unchanged.

\begin{figure}[!th]
\centering
    \includegraphics[width=0.38\textwidth]{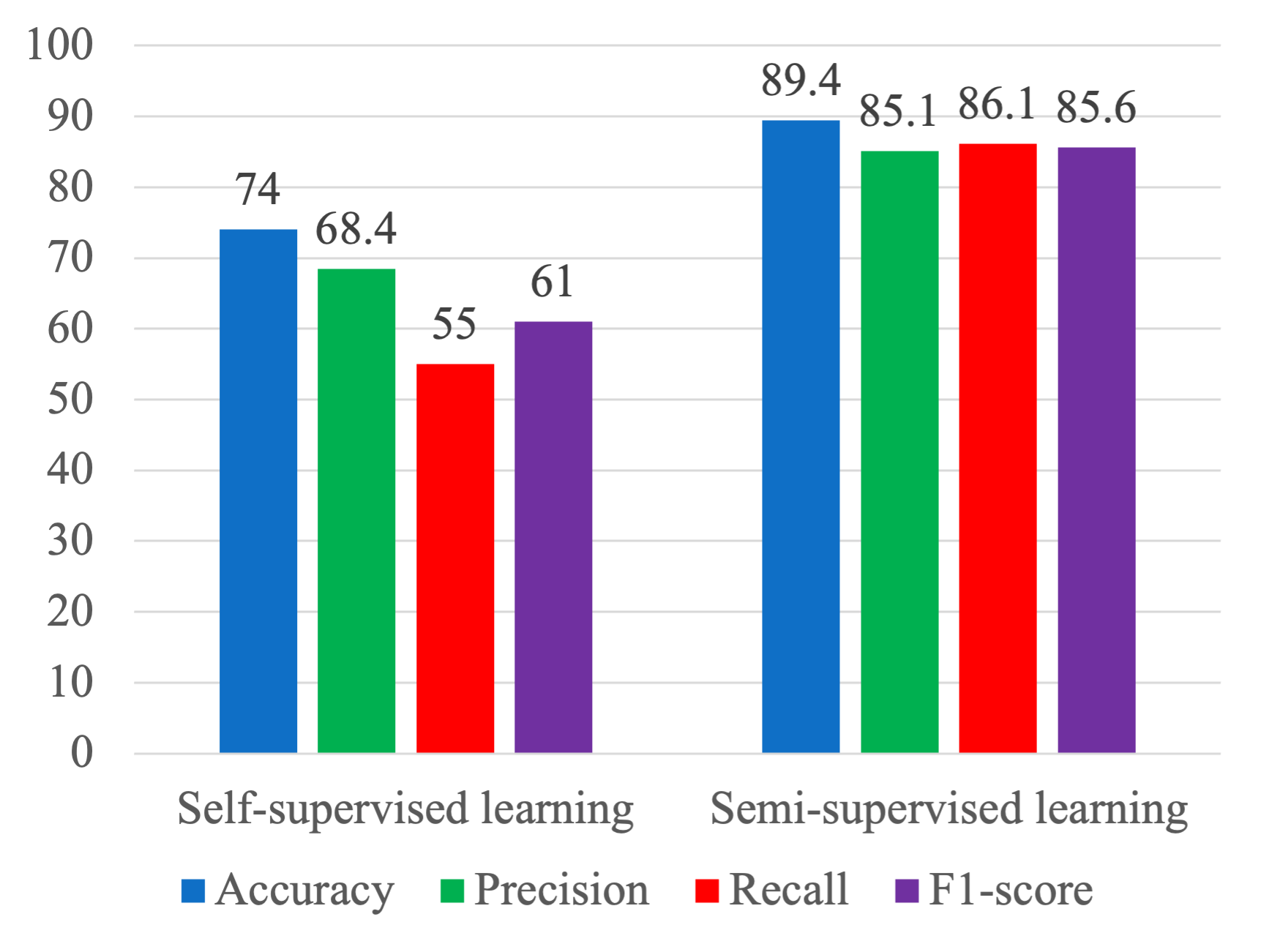}
    \vspace{-5mm}
  \caption{Semi-supervised learning in comparison to self-supervised learning.}
  \label{fig:shakya_2_semi_appendix}
\end{figure}

We utilize the reference genomes in the Shakya dataset to apply our semi-supervised learning method on Shakya 2, comparing it with the self-supervised approach.
As shown in Figure~\ref{fig:shakya_2_semi_appendix}, semi-supervised learning outperforms the self-supervised method in all of the metrics.
This experiment highlights two important features of our method. 
First, even though we present the method for the more challenging case where no information is given about the metagenomic sample other than the reads, whenever reference genomes are available, this information can be seamlessly introduced.
Second, this ground-truth information about the repetitive nature of some unitigs can drastically help in the repeat detection process, so it must be used whenever it is available.

\end{document}